\documentclass[11pt,a4paper]{article}
\usepackage[hyperref]{acl2021}
\usepackage{times}
\usepackage{latexsym}

\usepackage{microtype}

\aclfinalcopy 

\title{The Scenario Refin\emph{er}:

Grounding subjects in images at the morphological level}

\author{Tagliaferri Claudia \\
  \begin{tabular}{@{}c@{}}
  Utrecht University \\
  \texttt{\scriptsize cl.tagliaferri@outlook.com}
  \end{tabular} \\[1ex]
  \And
  Axioti Sofia \\
  \begin{tabular}{@{}c@{}}
  Leiden University \\
  \texttt{\scriptsize s.axioti@outlook.com}
  \end{tabular} \\[1ex]
  \And
  Gatt Albert \\
  Utrecht University \\
  \texttt{\scriptsize a.gatt@uu.nl} \\[1ex]
  \And
  Paperno Denis\\
  Utrecht University \\
  \texttt{\scriptsize d.paperno@uu.nl}}

\date{}

\usepackage{graphicx}
\usepackage{subfigure}

\begin{document}
\maketitle

\begin{abstract}
Derivationally related words, such as ``runner'' and ``running'', 
exhibit semantic differences which also elicit different visual scenarios. In this paper, we ask whether Vision and Language (V\&L) models capture such distinctions at the morphological level, using a a new methodology and dataset. We compare the results from V\&L models to human judgements and find that models' predictions differ from those of human participants, in particular displaying a grammatical bias. We further investigate whether the human-model misalignment is related to model architecture. Our methodology, developed on one specific morphological contrast, can be further extended for testing models on capturing other nuanced language features. 

\end{abstract}

\section{Introduction}

Vision and language (V\&L) models are trained to ground linguistic descriptions in visual data. 
These models differ in pre-training and architecture. 
In particular, there are differences in the cross-modal information exchange between the textual and visual streams of the models \cite{frank2021, parcalabescu_mm-shap_2022}, even though sometimes, as shown for V\&L models based on the BERT architecture \cite{devlin-etal-2019-bert}, architectural differences have little impact on downstream performance for many benchmarks \citep{bugliarello2021}.

Pre-trained V\&L models achieve high performance on diverse benchmarks, such as question answering, image retrieval and word masking \cite{tan2019}. However, they have limitations in tasks requiring \textit{fine-grained} understanding \cite{bugliarello2023measuring}, including the ability to reason compositionally in visually grounded settings \cite{thrush2022winoground}, distinguish spatial relationships and quantities \cite{parcalabescu2020,parcalabescu-etal-2022-valse}, and identify dependencies between verbs and arguments \cite{hendricks2021probing}. Most of these fine-grained linguistic phenomena are at the interface between syntax and semantics. 

Far less attention has been paid to grounding fine-grained linguistic features at the morphological level. We aim to address this gap by investigating multimodal alignment at the morphological level. We focus on derived nouns with the agentive suffix {\em -er} (e.g. {\em baker}) and the corresponding verbal form ({\em baking}). Such derivationally related pairs involve both category-level and semantic contrasts, with corresponding differences in the typical visual scenarios they evoke. For instance, human judges would accept the description {\em x is baking} for a variety of visual scenes depicting a person (hereafter referred to as `the subject') performing a particular action. Only a subset of such images would, however, also be judged as corresponding to {\em x is a baker}, since the agentive noun introduces additional expectations, for example about the way the subject is dressed or the physical environment they are in. By analysing the same stem (e.g.\ \textit{bake}) in different parts of speech, we explore the ability of V\&L models to capture the subtle differences in meaning and visual representation. 
To do this, we rely on a zero-shot setting in which we test the probability with which pretrained V\&L models match an image to a corresponding text containing an agentive noun or a verb, comparing this to human judgments about the same image-text pairs.

Our contributions are: (i) a methodology for testing V\&L models on morphological contrasts; (ii)  a dataset of images that highlights the contrast between verbs and derived nouns, annotated with human judgements; (iii) an analysis of the V\&L models' predictions on the contrast between derivationally related verbs and nouns, in comparison to human judgements.

\section{Related work}

\subsection{Models}
Various V\&L model architectures have been proposed, differing a.o.\ in the way visual vs.\ textual features are processed. One important distinction, common among models based on the BERT architecture, is between single- and dual-stream models. The former concatenate inputs in the two modalities and process them through a common transformer stack; the latter first process each modality through its own transformer stack, before performing cross-modal attention at a later stage \cite{bugliarello2021}. Another influential architecture is the dual encoder \cite{radford2021}, 
which is trained to project visual and textual embeddings into a common multimodal space. Among their pretraining objectives, BERT-based V\&L models typically include {\em image-text matching}, whereby the model returns a probability that an image corresponds with a caption. Thus, such models can be tested zero-shot on image-text pairs. For dual encoders, similar insights can be obtained by comparing the distance in multimodal space between a text and an image embedding.

We aim to understand the impact of these architectures on the morphological contrast between word categories and whether the classification depends on specific visual information. Three models with different architectures and pre-training phases are tested: CLIP \cite{radford2021}, ViLT \cite{kim2021}, and LXMERT \cite{tan2019}.

\paragraph{CLIP} employs a \emph{dual encoder} architecture and projects image and text embeddings in a common space, such that corresponding image-text pairs are closer than non-corresponding ones. CLIP is pre-trained using cross-modal contrastive learning on internet-sourced image-text pairs, resulting in strong multimodal representations \cite{radford2021}. Two different visual backbones are used for the image encoder: ResNet50 \cite{he_deep_2016}, which uses attention pooling; and the Vision Transformer \cite{Dosovitskiy2020} which is modified by the addition of an additional layer normalisation to the combined patch and position embedding. The text encoder is a Transformer which operates on a lower-cased byte pair encoding (BPE) representation of the text. CLIP computes the cosine similarity between an image and a text.

\paragraph{LXMERT} follows a \emph{dual-stream} approach, utilising three encoders: an object relationship encoder which acts upon the output of a faster-RCNN visual backbone \cite{Ren2015}, a language encoder, and a cross-modality transformer stack which applies attention across the two modalities. The pre-training involves five tasks, including masked cross-modality language modelling and image question answering, enabling the model to establish intra-modality and cross-modality relationships \cite{tan2019}. LXMERT is also pretrained with an image-text alignment head, which computes the probability that a text and an image correspond.

\paragraph{ViLT} \cite{kim2021} is the simplest V\&L architecture used in this study. It is a single-stream model in which a single transformer stack processes the concatenation of visual and textual features. In contrast to other models, no pre-trained visual backbone is used; rather, the model works directly on pixel-level inputs, in the spirit of \citet{Dosovitskiy2020}. 
It has been shown that the usage of word masking and image augmentations improves its performance \cite{kim2021}. In ViLT, the embedding layers of raw pixels and text tokens are shallow and computationally light. This architecture thereby concentrates most of the computation on modelling modality interactions.
Like LXMERT, ViLT is also pre-trained with an image-text alignment head, in addition to the multimodal masked modelling objective.

\subsection{Related studies}
Our work is related to studies focusing on the \emph{typicality} of the word-image relationship and the interplay with category labels for images depicting people. For example, people can be described using generic expressions referring to gender or more specific expressions highlighting individual properties or aspects. 
Visual properties that align with our conceptual knowledge of the noun may lead us to prefer agentive expressions over generic nouns such as  ``man'' or ``woman'' \cite{corbetta2021}. 
\citet{gualdoni2022a, gualdoni2022b} proposed ManyNames, a small dataset that explores the factors that affect naming variation for visual objects, for instance, the different conceptualisations of the same object (e.g., ``woman'' vs. ``tennis player'') or the disambiguation of the nature of the object (e.g., ``horse'' vs. ``pony''). Understanding the effects of context and naming preferences is crucial for V\&L models to gain comprehensive understanding of visual scenes. The \textit{typicality of the context} determines the occurrence of specific names based on the global scene where the subject is situated. 

The current study explores the impact of typicality of the context at the morphological level. 
Derivational relations, relating two words or whole paradigms of words \cite{bonami2019paradigm}, involve contrasts at different levels, including form, syntax -- where the words are related but belong to different word categories -- and semantics, where the meaning of one member contrasts with the meanings of the other members. For instance, \textit{runner} and \textit{run} belong to the same paradigm, but the suffix \textit{-er} changes the word category and alters the referential meaning of the verb. 
For example, ``the man is a runner'' evokes a fit person who frequently trains, while ``the man is running'' could equally well portray a man casually running to catch a train. Thus, derived noun subjects should embody characteristics of the verb and/or common knowledge. Therefore, syntactic and relational knowledge has to be integrated with semantic knowledge, common imaginary and visual information, as has been argued from the language acquisition perspective \cite{TYLER1989649}. 

\section{Methodology}

\subsection{Dataset}

\begin{table}
\small
\centering
\begin{tabular}{|cc||cc|}
\hline
{\bf Noun} & {\bf Verb} & {\bf Noun} & {\bf Verb}\\
\hline
supporter & supporting & lover & loving\\
baker & baking & surfer & surfing\\
runner & running & swimmer & swimming\\
hunter & hunting & driver & driving\\
painter & painting & skier & skiing\\
walker & walking & dancer & dancing\\
singer & singing & gamer & gaming\\
teacher & teaching & reader & reading\\
cleaner & cleaning & smoker & smoking\\
\hline
\end{tabular}
\caption{Noun-verb pairs in the Scenario Refin{\em er} dataset}\label{tab:lexical-items}
\end{table}

We create the Scenario Refin\emph{er} dataset highlighting the cognitive and semantic differences between the verb and its derived noun by contrasting one image with two annotations. The dataset is based on 18 word pairs, each consisting of a verb in the {\em -ing} form and a derived agentive ({\em -er}) noun. The pairs are summarised in Table \ref{tab:lexical-items}.
The lexical pairs are classified into four conceptual domains: the professional domain (like \emph{baker} or \emph{teacher}), the sports domain (like \emph{runner} or \emph{skier}), the artistic domain (like \emph{dancer} or \emph{painter}), and general  (\emph{lover} or \emph{smoker}). 

Six images were selected for each of the 18 word pairs. These were manually selected from various sources: Visual Genome \cite{krishna2017visual}, Wikipedia Commons, MSCOCO \cite{cocodataset} and Geograph (\url{https://www.geograph.org.uk/}).

For the 18 word pairs, we want to compare images which correspond to the stereotypical representation of the agent role described by the derived noun, versus the more general scenario described by the verb. In order to depict the subject denoted by a derived noun, the images need to include additional information compared to the verb, for example, specific objects like tools or outfits for \emph{painter} or \emph{surfer}; or a specific environment like a stage for \emph{dancer} or \emph{singer}. The verbs correspond to a more general scenario, which creates a linguistic and visual contrast with the scenario evoked by the derived noun. This allows us to examine the contrast in parts of speech and their typicality within the defined global scene \cite{gualdoni2022b}. 

\begin{figure}
    \centering
    \subfigure[]{\includegraphics[height=1.0in,width=1.0in]{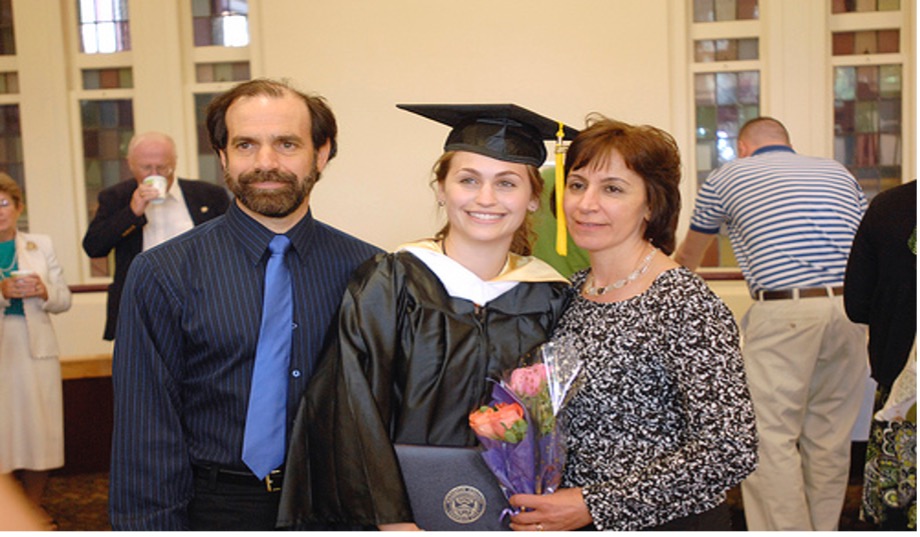}} 
    \caption*{Annotation 1: The man and the woman are supporters\\ Annotation 2: The man and the woman are supporting}
    {\includegraphics[height=1.0in,width=1.0in]{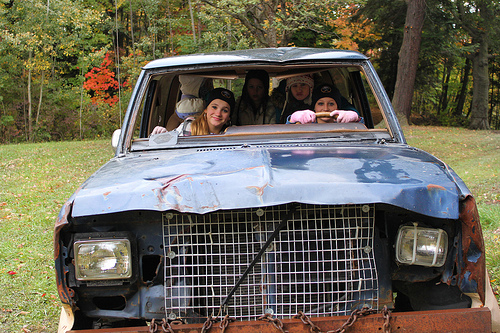}}
    \caption*{Annotation 1: The woman with pink gloves is a driver\\ Annotation 2: The woman with pink gloves is driving}
    \caption{Sample of stimuli for \textit{supporter-supporting} and \textit{driver-driving}}
    \label{fig:sample}
\end{figure}

For each word pair, 6 images were selected. Each image is accompanied by two captions,
as shown in Figure \ref{fig:sample}. Each caption received a judgement on a Likert scale.

\subsection{Data collection}

We implemented a survey on Qualtrics and distributed it on Prolific. The survey included 162 images, consisting of 54 fillers and (18 $\times$ 6 =) 108 target images representing the 18 selected lexical pairs. 

Our survey also included fillers of several types
. In one type, images were accompanied by a verb-based description and a derived noun in \emph{-er}, enhanced by an adjective based on the mood or facial expression of the depicted subjects. For instance, a smiling subject wearing appropriate outfit on a ski slope was paired with the captions ``The man is a \textit{happy skier}'' and ``The man \textit{is skiing}''. This type of filler aimed to investigate if participants would alter their evaluation when the mental representation of the derived noun is reinforced by additional linguistic information. Another type of filler contrasted the verb and its derived adjective in \emph{-ive}, offering insights into the classification of other members in the morphological paradigm. For example, four men intensely engaged in a video game were paired with the sentences ``The men are \textit{competitive}'' and ``The men \textit{are competing}''. A third type of filler contrasted verbs to bare adjectives, descriptive or emotional, to determine participants' preference between verbal and adjectival descriptions. For instance, a couple swimming happily in a lake was matched with ``The man and woman are happy'' and ``The man and woman are swimming''; an image of a man speaking in a classroom was paired with ``The man is upright'' and ``The man is teaching''. The fourth type of filler included images with true and false descriptions of the visual content, used to maintain participants' attention and allowing to control the quality of their responses. 

For each image, participants were asked to what extent both captions describe the visual scenario, using a seven-point Likert scale ranging from {\em totally disagree} to {\em totally agree}. By asking to evaluate both captions for each picture, it is possible to extract a reliable measure of contrast between the derived noun and the verb. 

In order not to risk rough human evaluations and minimise participant dropout rates due to the length of the survey, the target images were divided equally between two surveys (each with a total of 81 images where 54 were target images and 27 fillers). 

Twenty native British English speakers completed the online questionnaire and were randomly assigned to one of the two surveys. Thus, each image is evaluated by 10 participants for both captions. For the instructions see Appendix \ref{sec:appendix}

\section{Results}
Our analysis proceeds in two stages. We first consider the {\em category preference}: for an image with two captions (one with a derived noun and one with a verb), we ask whether human judges (resp. V\&L models) exhibit a preference for the noun or the verb with respect to a given image. We then compute correlations between the preferences exhibited by human judges and by models for the two categories.

\subsection{The word category preference}
To analyse which of the two captions is preferred for each image by human judges, we compare the average ratings of the annotations. For V\&L models, we consider the difference in probability estimated by a model's image-text matching head (in the case of ViLT and LXMERT) for the caption containing the noun or verb, or the difference in cosine distance between image and caption embeddings (in the case of CLIP). Note that we include results for three versions of CLIP, with different visual backbones. We use a Fisher test to determine whether there is a significant difference in category preference between human judges and V\&L models.

Table \ref{tab:classification} displays the proportion of times the derived noun or the verb was preferred by humans and by each of the models.

\begin{figure}
    \centering
    \subfigure[M = 5.50 (noun, verb), SD = 1.20 (noun, verb)]{\includegraphics[height=1.0in,width=1.5in]{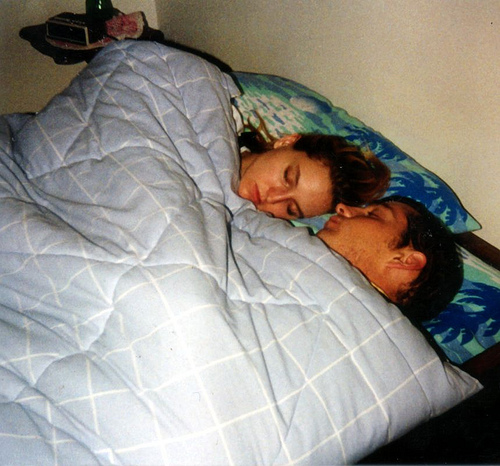}} 
    \subfigure[M = 6.30 (noun, verb), SD = 0.90 (noun, verb)]{\includegraphics[height=1.0in,width=1.5in]{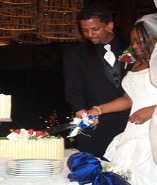}} 
    \subfigure[ M = 5.30 (noun, verb), SD = 0.90 (noun), 1.00 (verb)]{\includegraphics[height=1.0in,width=1.5in]{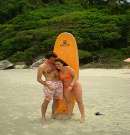}}
    \caption{Mean (M) human judgments and standard deviations (SD) for an example image set corresponding to {\em lover-loving}.}
    \label{fig:lover-loving}
\end{figure}

\paragraph{Human judgments} Overall, human judges exhibit a preference for captions containing the verb, with only a small percentage of preferences for captions containing agent nominals. These types of classifications are distributed across different domains. This could be due to variation in the images in the extent to which they gave clear visual cues as to the role of the person depicted. There were some exceptions to this trend. In the sports domain, these included images of a skier wearing skiing gear with a cape, and a couple of surfers in surfing attire with surfboards. In the profession domain, they included two images depicting individuals engaged in driving and one image of teachers with pupils posing for a class photo. Four agent nominals belonged to the artistic and general domains, such as images of women dancing on a stage, two subjects getting cigarettes, and a woman in a bookshop.
On the other hand, the difference in preference
some noun-verb pairs was lower than for others
(with differences in the 0--0.5 range). An example is shown in Figure \ref{fig:lover-loving}, where participants interpreted both captions as appropriate.
Interestingly, the versions of CLIP and LXMERT seem to agree with the human ratings in this example, showing low contrast between the verb and the noun, with LXMERT assigning higher probability to verb caption for (c) and CLIP estimating lower distance between image and verb caption for (a). On the other hand, ViLT assigned a higher probability to the verbal caption for all the images in Figure \ref{fig:lover-loving}.

\begin{figure}
    \centering
    \subfigure[noun: M = 6.20, SD = 1.17; verb: M = 6.50, SD = 1.02]{\includegraphics[height=1.0in,width=1.0in]{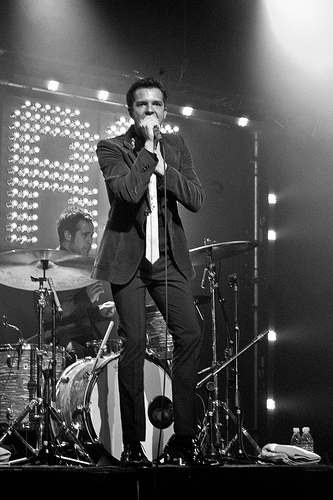}} 
    \subfigure[noun: M = 6.20, SD = 1.17; verb: M = 6.50, SD = 1.02]{\includegraphics[height=1.0in,width=1.0in]{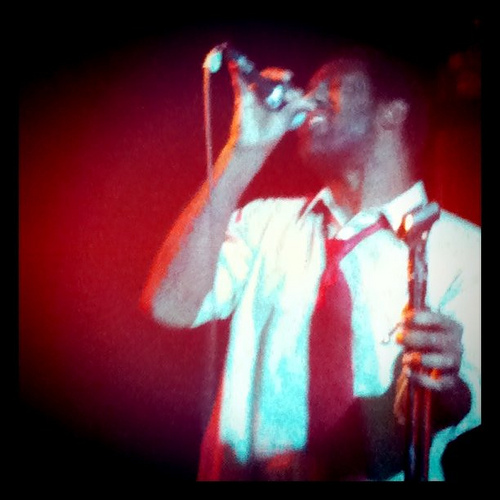}} 
    \caption{Mean (M) human judgments and standard deviations (SD) for an example image set corresponding to {\em singer-singing}}.
    \label{fig:singer-singing}
\end{figure}

\paragraph{V\&L models} Unlike participants, V\&L models exhibit a \textbf{tendency to prefer deverbal  nouns to verbs}. The exceptions are CLIP with the ViT-B/32 backbone, and ViLT, both of which have a slightly higher preference for captions with verbs. The performance of CLIP seems to depend on the visual backbone. Of the three versions, ViT-L/14 displays the greatest similarity to human judgments. We observed a tendency for ViT-B/32 to prefer captions with derived nouns where there are clear visual cues suggesting a role or activity, such as the microphone and the stage in Figure~\ref{fig:singer-singing}. In contrast, while CLIP-RN50 prefers the noun caption in Figure~\ref{fig:singer-singing}(a), it shows the opposite trend, in favour of the verb-based caption, in (b), perhaps because the stage is less clearly visible.

The difference between the judgements of humans vs.\ V\&L models is statistically significant (Fisher's exact test, p $<$ 0.001 for all contrasts between models and human judgments). 

\begin{table}
\centering
\small
\begin{tabular}{lrl}
\hline \textbf{ } &\textbf{Derived noun} & \textbf{Verb} \\ \hline
Humans & 8.3\% & 91.7\% \\
CLIP ViT-L/14@336px & 51.9\% & 48.1\% \\
CLIP RN50x64 & 52.8\% & 47.2\% \\
CLIP ViT-B/32 & 49.1\% & 50.9\%  \\
ViLT & 47.2\% & 52.8\% \\
LXMERT & 51.9\% & 48.1\%  \\
\hline
\end{tabular}
\caption{\label{tab:classification} Preference for derived noun vs. verb, in human judgments and V\&L model image-text alignment.}
\end{table}

\subsection{Correlations between judgements}

We also estimate the correlation between human and automatic judgements as a more fine-grained measure than binary preference. Overall, the correlation between the human and the automatic judgements varies depending on architecture and on the conceptual domain. 

We assess correlations between three kinds of values: the (human- or model-produced) scores for a)~noun and b)~verb-based captions, as well as c)~the difference between the noun and verb scores. We refer to the latter as the {\em morphological contrast}.

\paragraph{Participant consistency} To assess the consistency of collected human judgements, we split participants randomly into two equal-sized samples and calculate Pearson correlation coefficients between the average scores of the two samples.
The resulting correlation coefficients for all conceptual domains 
are reported in Table \ref{tab:participants}. Correlation coefficients for noun, verb and contrast are generally consistent, with the exception of the artistic domain, for which correlations between judgments for verb-based captions, and as a consequence, also for the contrast, exhibit more variation.


\begin{table*}
\centering
\begin{tabular}{lccc}
\hline
\textbf{Domain} &\textbf{Derived noun} &\textbf{Verb} &\textbf{Morphological contrast} \\
\hline
Professional domain & 0.76 & 0.84 & 0.75 \\
Sport domain & 0.69 & 0.70 & 0.60 \\
Artistic domain & 0.79 & 0.31 & 0.51\\
General & 0.92 & 0.88 & 0.94 \\
All domains & 0.80 & 0.81 & 0.78 \\
\hline
\end{tabular}
\caption{\label{tab:participants} Human judgements: Pearson correlations of judgments for captions containing derived nouns and verbs, and for the difference (contrast).}
\end{table*}

\begin{table*}
\centering
\begin{tabular}{lccc}
\hline
\textbf{Model} &\textbf{Derived noun} &\textbf{Verb} &\textbf{Morphological contrast} \\
\hline
CLIP ViT-L/14@336px & 0.13 & 0.08 & 0.15 \\
CLIP RN50x64 & 0.09 & 0.08 & -0.01 \\
CLIP ViT-B/32 & 0.09 & 0.18 & 0.08 \\
ViLT & 0.07 & 0.26 & 0.32 \\
LXMERT & 0.16 & 0.03 & 0.21 \\
\hline
\end{tabular}
\caption{\label{tab:models-general} Human judgments and V\&L models overall: Pearson correlations between human judgements and model image-text alignment for captions containing derived nouns, verbs, and the contrast between them.}
\end{table*}

\begin{table*}[ht]
\centering
\small
\begin{tabular}{lccc|ccc}
\hline
\multicolumn{1}{c}{} & \multicolumn{3}{c}{\textbf{Sport domain}} & \multicolumn{3}{c}{\textbf{Professional domain}} \\
\multicolumn{1}{c}{} & \textbf{Deriv. noun} & \textbf{Verb} & \textbf{Morph. contrast} & \textbf{Deriv. noun} & \textbf{Verb} & \textbf{Morph. contrast} \\
\hline
CLIP ViT-L/14@336p & 0.02 & -0.25 & -0.06 & -0.04 & 0.19 & 0.33 \\
CLIP RN50x64 & 0.02 & -0.31 & -0.11 & -0.22 & 0.33 & 0.23 \\
CLIP ViT-B/32 & -0.04 & -0.26 & -0.11 & -0.16 & 0.23 & 0.40 \\
ViLT & -0.01 & -0.04 & 0.45 & 0.30 & 0.45 & 0.68 \\
LXMERT & 0.08 & -0.32 & 0.17 & -0.10 & 0.18 & 0.40 \\
\hline
\multicolumn{1}{c}{} & \multicolumn{3}{c}{\textbf{Artistic domain}} & \multicolumn{3}{c}{\textbf{General}} \\
\multicolumn{1}{c}{} & \textbf{Derived noun} & \textbf{Verb} & \textbf{Contrast} & \textbf{Derived noun} & \textbf{Verb} & \textbf{Contrast} \\
\hline
CLIP ViT-L/14@336p & -0.03 & 0.01 & 0.22 & 0.28 & 0.18 & -0.06 \\
CLIP RN50x64 & 0.06 & 0.21 & 0.27 & 0.08 & 0.23 & 0.10 \\
CLIP ViT-B/32 & -0.09 & -0.04 & -0.007 & 0.23 & 0.25 & -0.06 \\
ViLT & 0.44 & 0.15 & 0.39 & -0.06 & 0.05 & 0.25 \\
LXMERT & 0.29 & 0.26 & 0.42 & -0.01 & -0.25 & 0.12 \\
\hline
\end{tabular}
\caption{\label{tab:model-domains} Human judgement and V\&L models by domain: Pearson correlation between human judgments and model image-text matching estimates for captions containing derived nouns, verbs, and the morphological contrast between the derived noun and the verb.}
\end{table*}

\paragraph{Models vs human judgments} Table~\ref{tab:models-general} displays the overall correlations between human judgments and model image-text alignment for verbs, nouns and the morphological contrast. The correlations are moderate-to-weak, suggesting a lack of alignment between human intuitions and V\&L models. This is consistent with our earlier observation that models tend to exhibit different preferences for nouns versus verbs, compared to humans. Interestingly, ViLT emerges as the most correlated model with human judgement in the verbal evaluation, but it exhibits the least correlation in the evaluation of the derived noun. Additionally, ViLT displays a moderate positive relationship with the contrast between verb and derived noun, whereas the other models demonstrate weaker positive correlations or very weak negative correlations with this particular contrast.


Table~\ref{tab:model-domains} breaks down correlations by conceptual domain. In the professional domain, correlations are generally stronger, especially for ViLT, LXMERT and CLIP ViT-B/32. Overall, it appears that models correlate with human judges in some domains more than others. Nevertheless, correlations are often negative, and these results suggest a qualitative difference between the image-text alignment performed by models, and the types of knowledge and inferences that humans bring to bear to support the grounding of nominal agentive versus verbal forms in visual stimuli.

\section{Discussion}

The findings revealed a discrepancy between models and human judgments. Humans displayed a preference for captions containing verbs, whereas V\&L models exhibited a preference for nominal descriptions. Participants prefer the derived noun only for a few instances that had additional characteristics elicited by visual elements, or by the kind of action performed by the human subjects in the images. For instance, they prefer the derived noun for two images showing a person getting or purchasing cigarettes (\textit{smoker-smoking}), meaning that participants interpreted the \textit{intention} as a characteristic that corresponds to the derived noun. In contrast, the tested models appeared to prioritise more the action itself rather than the individual who performs the action. 

However, examining certain lexical pairs, we observed a greater variance in the pattern of interpretation, highlighting the difficulty in defining the human evaluation of the derived noun. For example, in the sport domain, participants rarely seem to rely on the outfit worn by the subject to base their interpretation, with the exception of \emph{skier}, which happened to be paired only with an image of a subject also exhibiting their competition number. As a surprising contrast, two pictures for \emph{runner-running} similarly depicted subjects with their competition numbers are not evaluated as such by participants. Specifically, one image depicts a man running in a race track, while the other image depicts three men wearing specific outfits running in the countryside. The contrast between the means of the human evaluation is less than or equal to 0.50, indicating the preference for the verbal description. 

The models, too, exhibit variety in the subject classification for these images. For example, while CLIP-ViT-L/14@336p, CLIP-ViT-B/32 and ViLT display a similar preference for the nominal form, as humans do, for \emph{skier-skiing}, CLIP-RN50x64 and LXMERT prefer the verb-based caption. Similarly, while participants slightly prefer the verb for the subjects wearing a competition number for \emph{runner-running}, models prefer the nominal description. The three versions of CLIP strongly prefer the derived noun for these subjects, ViLT prefers the verbal description only for the single subject running in a race track and LXMERT prefers the verbal description only for the three subjects running in the countryside. While CLIP exhibited a preference for the derived noun in presence of additional visual elements, ViLT and LXMERT do not seem to base their preference on such a visual cue since they assign a high probability to the verbal description too. 


\section{Conclusion}

We studied the morphological difference between derived nouns in \emph{-er} and verbs for visual grounding, comparing human judgements with pre-trained Vision and Language models. The dataset we presented allows us to assess vision and language models on their understanding of verbs, deverbal agent nouns, and most importantly the contrast between the two. Our results show that while some models, especially ViLT, show strong results for some of the conceptual domains, they do not support the conclusion that models ground the morphological differences between derived nouns and verbs in a humanlike way. 

Highlighting and investigating such a morphological and cognitive difference can refine and improve the alignment of textual and visual input of V\&L models. By exploring the visual classification at the morphological level, the aim was to investigate not only the linguistic and morphological influence in the automatic recognition of subjects carrying certain visual information, but also to individuate which architecture of the model better executes the task. In our study, the single-stream ViLT model tends to correlate better with human judgments. Nevertheless, these results are based on a relatively small test set and focus on a restricted set of models, with much scope for further experimentation. In an effort to encourage the community to undertake further investigation of these phenomena, we have shared our code and our dataset text.\footnote{https://github.com/ClaudiaTagliaferri/Scenario\_Refiner.git}.

\bibliographystyle{acl_natbib}
\bibliography{anthology,acl2021}

\begin{thebibliography}{22}
\expandafter\ifx\csname natexlab\endcsname\relax\def\natexlab#1{#1}\fi

\bibitem[{Bonami and Strnadov{\'a}(2019)}]{bonami2019paradigm}
Olivier Bonami and Jana Strnadov{\'a}. 2019.
\newblock Paradigm structure and predictability in derivational morphology.
\newblock \emph{Morphology}, 29(2):167--197.

\bibitem[{Bugliarello et~al.(2021)Bugliarello, Cotterell, Okazaki, and
  Elliott}]{bugliarello2021}
Emanuele Bugliarello, Ryan Cotterell, Naoaki Okazaki, and Desmond Elliott.
  2021.
\newblock Multimodal pretraining unmasked: A meta-analysis and a unified
  framework of vision-and-language berts.
\newblock \emph{Transactions of the Association for Computational Linguistics},
  9:978--994.

\bibitem[{Bugliarello et~al.(2023)Bugliarello, Sartran, Agrawal, Hendricks, and
  Nematzadeh}]{bugliarello2023measuring}
Emanuele Bugliarello, Laurent Sartran, Aishwarya Agrawal, Lisa~Anne Hendricks,
  and Aida Nematzadeh. 2023.
\newblock Measuring progress in fine-grained vision-and-language understanding.
\newblock \emph{arXiv preprint arXiv:2305.07558}.

\bibitem[{Corbetta(2021)}]{corbetta2021}
Daniela Corbetta. 2021.
\newblock Effects of typicality and category label on referring expressions in
  context: Empirical analysis in the domain of “people”.
\newblock \emph{Departament de Traducció i Ciències del Llenguatge}.

\bibitem[{Devlin et~al.(2019)Devlin, Chang, Lee, and
  Toutanova}]{devlin-etal-2019-bert}
Jacob Devlin, Ming-Wei Chang, Kenton Lee, and Kristina Toutanova. 2019.
\newblock \href {https://doi.org/10.18653/v1/N19-1423} {{BERT}: Pre-training of
  deep bidirectional transformers for language understanding}.
\newblock In \emph{Proceedings of the 2019 Conference of the North {A}merican
  Chapter of the Association for Computational Linguistics: Human Language
  Technologies, Volume 1 (Long and Short Papers)}, pages 4171--4186,
  Minneapolis, Minnesota. Association for Computational Linguistics.

\bibitem[{Dosovitskiy et~al.(2020)Dosovitskiy, Beye, Kolesnikov, Weissenborn,
  Zhai, Unterthiner, Dehghani, Minderer, Heigold, Gelly, Uszkoreit, and
  Houlsby}]{Dosovitskiy2020}
Alexey Dosovitskiy, Lucas Beye, Alexander Kolesnikov, Dirk Weissenborn, Xiaohua
  Zhai, Thomas Unterthiner, Mostafa Dehghani, Matthias Minderer, Georg Heigold,
  Sylvain Gelly, Jakob Uszkoreit, and Neil Houlsby. 2020.
\newblock An image is worth 16x16 words: {Transformers} for image recognition
  at scale.
\newblock \emph{arXiv}, 2010.11929.

\bibitem[{Frank et~al.(2021)Frank, Bugliarello, and Elliott}]{frank2021}
Stella Frank, Emanuele Bugliarello, and Desmond Elliott. 2021.
\newblock Vision-and-language or vision-for-language? on cross-modal influence
  in multimodal transformers.
\newblock \emph{arXiv preprint arXiv:2109.04448}.

\bibitem[{Gualdoni et~al.(2022{\natexlab{a}})Gualdoni, Brochhagen,
  M{\"a}debach, and Boleda}]{gualdoni2022a}
Eleonora Gualdoni, Thomas Brochhagen, Andreas M{\"a}debach, and Gemma Boleda.
  2022{\natexlab{a}}.
\newblock Horse or pony? visual typicality and lexical frequency affect
  variability in object naming.
\newblock \emph{Proceedings of the Society for Computation in Linguistics},
  5(1):241--243.

\bibitem[{Gualdoni et~al.(2022{\natexlab{b}})Gualdoni, Brochhagen,
  M{\"a}debach, and Boleda}]{gualdoni2022b}
Eleonora Gualdoni, Thomas Brochhagen, Andreas M{\"a}debach, and Gemma Boleda.
  2022{\natexlab{b}}.
\newblock Woman or tennis player? visual typicality and lexical frequency
  affect variation in object naming.
\newblock In \emph{Proceedings of the Annual Meeting of the Cognitive Science
  Society}, volume~44.

\bibitem[{He et~al.(2016)He, Zhang, Ren, and Sun}]{he_deep_2016}
Kaiming He, Xiangyu Zhang, Shaoqing Ren, and Jian Sun. 2016.
\newblock \href {https://doi.org/10.1109/CVPR.2016.90} {Deep {Residual}
  {Learning} for {Image} {Recognition}}.
\newblock In \emph{2016 {IEEE} {Conference} on {Computer} {Vision} and
  {Pattern} {Recognition} ({CVPR})}, pages 770--778, Las Vegas, NV, USA. IEEE.

\bibitem[{Hendricks and Nematzadeh(2021)}]{hendricks2021probing}
Lisa~Anne Hendricks and Aida Nematzadeh. 2021.
\newblock Probing image-language transformers for verb understanding.
\newblock \emph{arXiv preprint arXiv:2106.09141}.

\bibitem[{Kim et~al.(2021)Kim, Son, and Kim}]{kim2021}
Wonjae Kim, Bokyung Son, and Ildoo Kim. 2021.
\newblock Vilt: Vision-and-language transformer without convolution or region
  supervision.
\newblock In \emph{International Conference on Machine Learning}, pages
  5583--5594. PMLR.

\bibitem[{Krishna et~al.(2017)Krishna, Zhu, Groth, Johnson, Hata, Kravitz,
  Chen, Kalantidis, Li, Shamma et~al.}]{krishna2017visual}
Ranjay Krishna, Yuke Zhu, Oliver Groth, Justin Johnson, Kenji Hata, Joshua
  Kravitz, Stephanie Chen, Yannis Kalantidis, Li-Jia Li, David~A Shamma, et~al.
  2017.
\newblock Visual genome: Connecting language and vision using crowdsourced
  dense image annotations.
\newblock \emph{International journal of computer vision}, 123:32--73.

\bibitem[{Lin et~al.(2014)Lin, Maire, Belongie, Bourdev, Girshick, Hays,
  Perona, Ramanan, Doll{'{a} }r, and Zitnick}]{cocodataset}
Tsung{-}Yi Lin, Michael Maire, Serge~J. Belongie, Lubomir~D. Bourdev, Ross~B.
  Girshick, James Hays, Pietro Perona, Deva Ramanan, Piotr Doll{'{a} }r, and
  C.~Lawrence Zitnick. 2014.
\newblock \href {http://arxiv.org/abs/1405.0312} {Microsoft {COCO:} common
  objects in context}.
\newblock \emph{CoRR}, abs/1405.0312.

\bibitem[{Parcalabescu et~al.(2022)Parcalabescu, Cafagna, Muradjan, Frank,
  Calixto, and Gatt}]{parcalabescu-etal-2022-valse}
Letitia Parcalabescu, Michele Cafagna, Lilitta Muradjan, Anette Frank, Iacer
  Calixto, and Albert Gatt. 2022.
\newblock \href {https://doi.org/10.18653/v1/2022.acl-long.567} {{VALSE}: A
  task-independent benchmark for vision and language models centered on
  linguistic phenomena}.
\newblock In \emph{Proceedings of the 60th Annual Meeting of the Association
  for Computational Linguistics (Volume 1: Long Papers)}, pages 8253--8280,
  Dublin, Ireland. Association for Computational Linguistics.

\bibitem[{Parcalabescu and Frank(2022)}]{parcalabescu_mm-shap_2022}
Letitia Parcalabescu and Anette Frank. 2022.
\newblock \href {https://doi.org/10.48550/arXiv.2212.08158} {{MM}-{SHAP}: {A}
  {Performance}-agnostic {Metric} for {Measuring} {Multimodal} {Contributions}
  in {Vision} and {Language} {Models} \& {Tasks}}.
\newblock ArXiv:2212.08158 [cs].

\bibitem[{Parcalabescu et~al.(2020)Parcalabescu, Gatt, Frank, and
  Calixto}]{parcalabescu2020}
Letitia Parcalabescu, Albert Gatt, Anette Frank, and Iacer Calixto. 2020.
\newblock Seeing past words: Testing the cross-modal capabilities of pretrained
  v\&l models on counting tasks.
\newblock \emph{arXiv preprint arXiv:2012.12352}.

\bibitem[{Radford et~al.(2021)Radford, Kim, Hallacy, Ramesh, Goh, Agarwal,
  Sastry, Askell, Mishkin, Clark et~al.}]{radford2021}
Alec Radford, Jong~Wook Kim, Chris Hallacy, Aditya Ramesh, Gabriel Goh,
  Sandhini Agarwal, Girish Sastry, Amanda Askell, Pamela Mishkin, Jack Clark,
  et~al. 2021.
\newblock Learning transferable visual models from natural language
  supervision.
\newblock In \emph{International conference on machine learning}, pages
  8748--8763. PMLR.

\bibitem[{Ren et~al.(2015)Ren, He, Girshick, and Sun}]{Ren2015}
Shaoqing Ren, Kaiming He, Ross Girshick, and Jian Sun. 2015.
\newblock Faster {R}-{CNN}: {Towards} {Real}-{Time} {Object} {Detection} with
  {Region} {Proposal} {Networks}.
\newblock In \emph{Advances in {Neural} {Information} {Processing} {Systems} 28
  ({NeurIPS} 2015)}, Montreal, Canada.

\bibitem[{Tan and Bansal(2019)}]{tan2019}
Hao Tan and Mohit Bansal. 2019.
\newblock Lxmert: Learning cross-modality encoder representations from
  transformers.
\newblock \emph{arXiv preprint arXiv:1908.07490}.

\bibitem[{Thrush et~al.(2022)Thrush, Jiang, Bartolo, Singh, Williams, Kiela,
  and Ross}]{thrush2022winoground}
Tristan Thrush, Ryan Jiang, Max Bartolo, Amanpreet Singh, Adina Williams, Douwe
  Kiela, and Candace Ross. 2022.
\newblock Winoground: Probing vision and language models for visio-linguistic
  compositionality.
\newblock In \emph{Proceedings of the IEEE/CVF Conference on Computer Vision
  and Pattern Recognition}, pages 5238--5248.

\bibitem[{Tyler and Nagy(1989)}]{TYLER1989649}
Andrea Tyler and William Nagy. 1989.
\newblock \href {https://doi.org/https://doi.org/10.1016/0749-596X(89)90002-8}
  {The acquisition of english derivational morphology}.
\newblock \emph{Journal of Memory and Language}, 28(6):649--667.

\end{thebibliography}

\appendix
\section{Appendix}
\label{sec:appendix}

Instructions for participants:
\par
Welcome to our survey! Our project focuses on improving existing annotation accompanying pictures. You will be presented with pictures and asked to indicate to which degree you agree with some statements. The study should take you around 15-20 minutes to complete. Your participation in this research will be paid only if you complete the survey. Please make sure to be redirected to Prolific at the end of the survey. In such a way, we can check if you completed the study and pay your participation.The ProlificID and all the sensitive data will be deleted once the payment is done. In the next page, you will be able to read more about the study and how we are doing with the data. If you would like to contact us to receive more information about the annotation project, please c.tagliaferri1@students.uu.nl or d.paperno@uu.nl.

\end{document}